\pdfoutput=1

\documentclass[10pt,twocolumn,letterpaper]{article}

\usepackage{cvpr}      % To produce the REVIEW version
\usepackage{graphicx}
\usepackage{color}

\usepackage{amsmath}
\usepackage{amssymb}
\usepackage{booktabs}
\usepackage{subcaption}

\usepackage[acronym,shortcuts]{glossaries}
\newacronym{DCNN}{DCNN}{deep convolutional neural networks}
\newacronym{IT}{IT}{inferotemporal cortex}
\newacronym{LGN}{LGN}{lateral geniculate nucleus}
\newacronym{LTD}{LTD}{long-term depression}
\newacronym{LTP}{LTP}{long-term potentiation}
\newacronym{RGC}{RGC}{retinal ganglion cell}
\newacronym{SNN}{SNN}{spiking neural network}
\newacronym{STDP}{STDP}{spike-timing-dependent-plasticity}
\newacronym{WTAI}{WTA-I}{winner-take-all inhibition}
\usepackage[pagebackref,breaklinks,colorlinks]{hyperref}
\usepackage{amsmath}
\usepackage{subcaption}

\usepackage[capitalize]{cleveref}
\crefname{section}{Sec.}{Secs.}
\Crefname{section}{Section}{Sections}
\Crefname{table}{Table}{Tables}
\crefname{table}{Tab.}{Tabs.}

\def\cvprPaperID{*****} % *** Enter the CVPR Paper ID here
\def\confName{CVPR}
\def\confYear{2022}

\begin{document}

%\title{\LaTeX\ Author Guidelines for \confName~Proceedings}
\title{Efficient visual object representation using a biologically plausible spike-latency code and winner-take-all inhibition}

% Important keywords that may or may not fit in the title:
% biologically plausible
% unsupervised
% low power, efficient
% using spike-latency encoding
% WTA

\author{Melani Sanchez-Garcia\\
\footnotesize Department of Computer Science \\ \footnotesize University of California, Santa Barbara, CA, USA\\
% {\tt\small mesangar@ucsb.edu}
\and
Tushar Chauhan\\
\footnotesize The Picower Institute for Learning and Memory, \\ \footnotesize Department of Brain and Cognitive Sciences, \footnotesize Massachusetts Institute of Technology \\
\footnotesize Centre de Recherche Cerveau et Cognition, \footnotesize Universit\'{e} de Toulouse, France \\
% {\tt\small tushar.chauhan@cnrs.fr}
\and
Benoit R. Cottereau \\
\footnotesize Centre de Recherche Cerveau et Cognition, \\ \footnotesize Universit\'{e} de Toulouse, France \\
% {\tt\small benoit.cottereau@cnrs.fr}
\and
Michael Beyeler\\
\footnotesize Department of Computer Science \\ \footnotesize Department of Psychological \& Brain Sciences \\ \footnotesize University of California, Santa Barbara, CA, USA\\
% {\tt\small mbeyeler@ucsb.edu}
}
\maketitle

%%%%%%%%% ABSTRACT
\begin{abstract}

\noindent Deep neural networks have surpassed human performance in key visual challenges such as object recognition, but require a large amount of energy, computation, and memory.
In contrast, \acfp{SNN} have the potential to improve both the efficiency and biological plausibility of object recognition systems.
Here we present a \acs{SNN} model that uses spike-latency coding and \acf{WTAI} to efficiently represent visual stimuli from the Fashion MNIST dataset.
Stimuli were preprocessed with center-surround receptive fields and then fed to a layer of spiking neurons whose synaptic weights were updated using \acf{STDP}.
We investigate how the quality of the represented objects changes under different \acs{WTAI} schemes and demonstrate that a network of 150 spiking neurons can efficiently represent objects with as little as 40 spikes.
% Studying how object recognition may be implemented
% using biologically plausible learning rules in \acsp{SNN} may not only further our understanding of the brain, but also lead to novel and efficient artificial vision systems.

% Brains perform very complex tasks consuming very low energy through neurons. However, designing computational models which use biologically plausible mechanisms especially for learning new patterns remains a challenging task. Recently, there is a growing interest in how \acf{SNN} can be used to perform complex computations or solve pattern representation tasks. This paper proposes a brain-inspired \acs{SNN} approach that extracts high informative visual features of a stimuli for an object representation task. Our model is trained with simulated neurons in primary layers that fire asynchronously, with the most strongly activated neurons firing first. We also simulated neurons at later stages of the system, called V1 neurons, implementing a \acf{STDP} rule and variations of \acf{WTAI} inhibition schemes. Our model shows that when the network is presented with only few stimuli images through a single \acs{SNN} layer, V1 neurons become selective to visual features that are present in the input image allowing object representation. These results also show that temporal codes and \acs{STDP} may be a key to understanding the processing in the visual system and can lead to fast and selective neural responses allowing object representation with few spikes.
\end{abstract}

%%%%%%%%% BODY TEXT
\section{Introduction}
\label{sec:intro}

\noindent Deep convolutional neural networks (DCNNs) have been extremely successful in a wide range of computer vision applications, rivaling or exceeding human benchmark performance in key visual challenges such as object recognition \cite{he2015}.
However, state-of-the-art DCNNs require too much energy, computation, and memory to be deployed on most computing devices and embedded systems \cite{goel2020survey}.
In contrast, the brain is masterful at representing real-world objects with a cascade of reflexive, largely feedforward computations \cite{dicarlo2012does} that rapidly unfold over time \cite{cichy2016} and rely on an extremely sparse, efficient neural code (see \cite{beyeler2019neural} for a recent review).
For example, faces are processed in localized patches within \ac{IT}, where cells detect distinct constellations of face parts (e.g., eyes, noses, mouths), and whole faces can be recognized by taking a linear combination of neuronal activity across \ac{IT} \cite{beyeler2019neural}.

\begin{figure*}[t]
  \centering
   \includegraphics[width=0.9\linewidth]{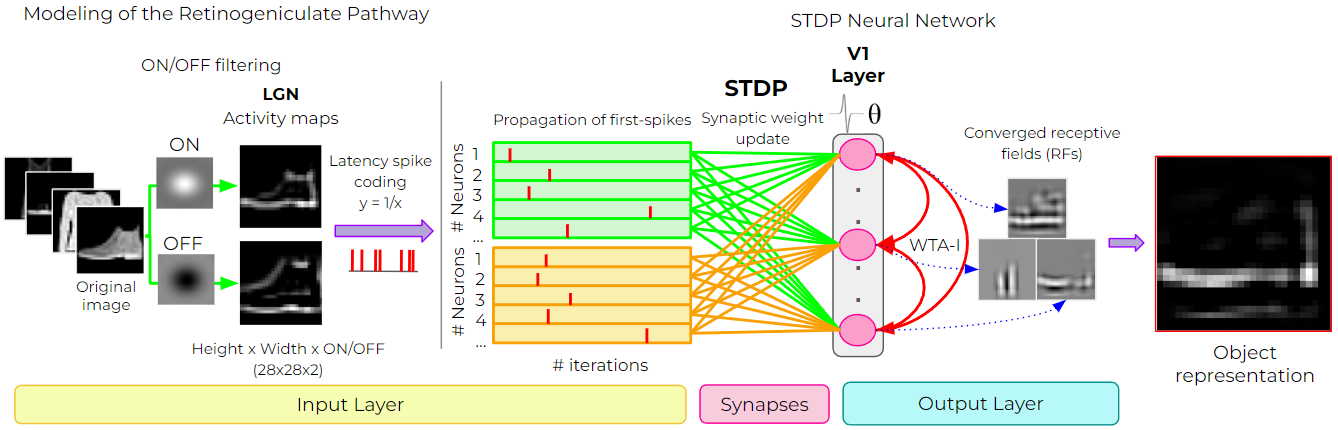}
   \caption{\textbf{Network architecture}. Images from the Fashion MNIST dataset were convolved with ON and OFF center-surround kernels to simulate responses in the \acf{LGN}. \Acs{LGN} responses were converted to spike latencies and fed to a \acf{SNN} with plastic synapses implementing \acf{STDP} and \acf{WTAI}. The propagated LGN spikes contributed to an increase in the membrane potential of V1 neurons until one of the V1 membrane potentials reached threshold, resulting in a postsynaptic spike and inhibition of all other V1 neurons until the next iteration. The synaptic weights were updated using an unsupervised \acs{STDP} rule.
   This allowed us to represent objects using the $\approx 40$ most active V1 neurons. }
   \label{fig:approach}
\end{figure*}

In recent years, \acp{SNN} have emerged as a promising approach to improving the efficiency and biological plausibility of neural networks such DCNN, due to their potential for low power consumption, fast inference, event-driven processing, and asynchronous operation.
Studying how object recognition may be implemented using biologically plausible learning rules in \acp{SNN} may not only further our understanding of the brain, but also lead to new efficient artificial vision systems.
% \Acp{SNN} have the ability to learn visual features in a biological manner with spiking neurons and unsupervised way using a technique called Spike-Timing-Dependent-Plasticity (\ac{STDP}) \cite{song2000competitive} which relies on sparsely encoded spiking information among local neurons. Furthermore, computational models that involve competitive stages are widely used in computational brain models. The simplest competitive computational module is the hard \ac{WTAI} scheme \cite{maass2000computational}.

Here we present a \ac{SNN} model that uses spike-latency encoding \cite{chauhan2018emergence} and \ac{WTAI} \cite{maass2000computational} to efficiently represent stimuli from the Fashion MNIST dataset \cite{xiao2017fashion}.
We show that efficient object representations can be learned with \ac{STDP} \cite{gutig_learning_2003}, an unsupervised learning rule that relies on sparsely encoded visual information among local neurons. In addition, we investigate how the quality of the represented objects changes under different \ac{WTAI} schemes.
Remarkably, our network is able to represent objects with as little as 150 spiking neurons and at most 40 spikes.

% --------------------------------

\section{Methods}
\label{section:methods}

% \noindent In this section, we describe our SNN model to efficiently represent stimuli from the Fashion MNIST dataset. After introducing some details about the network architecture, we describe the spike-latency encoding, STDP, and the WTA-I schemes.
% The model uses the fashion MNIST dataset for stimulus reconstruction, the latest being explained in Section \ref{section:methods}. These terms are explained in detail below.

\subsection{Network architecture}

\noindent The network architecture of our model is shown in  Figure~\ref{fig:approach}.
Inspired by \cite{chauhan2018emergence}, our network consisted of an input layer corresponding to a simplified model of the \ac{LGN}, followed by a layer of spiking neurons whose synaptic weights were updated using \ac{STDP}.
The \ac{LGN} layer consisted of simulated firing-rate neurons with center-surround receptive fields, implemented using a  6x6 difference of Gaussian filter (see Figure~\ref{fig:approach}, left).
The \ac{LGN} layer was fully connected to a layer of integrate-and-fire neurons, each unit characterized by a threshold and a membrane potential \cite{chauhan2018emergence}. Thus, the LGN spikes contributed to an increase in the membrane potential of V1 neurons, until one of the V1 membrane potentials reached threshold, resulting in a postsynaptic spike. The membrane potential $E_{n}(t)$ of the $n^{th}$ V1 neuron at time $t$ within the iteration was represented as:
\begin{equation}
\resizebox{0.9\hsize}{!}{
${E_{n}(t) =\begin{cases}  \sum\limits_{m\in LGN} w_{mn}\cdot H(t-t_{m}), \,\,\,\, & t < \min\limits_{t} \Big\{ t \mid \max\limits_{n \in V1} E_{n}(t) \ge \theta \Big\}  \\
          &         0, \,\,\,\, \text{otherwise}. \end{cases}}$}
    \label{eq:1.1}
\end{equation}
where $t_{m}$ was the spiking time of the $m$-th LGN neuron, ${H}$ was the Heaviside or unit step function, and $\theta$ was the threshold of the V1 neurons (assumed to be a constant for the entire population). The expression $\min \{ t \mid \max E_{n}(t) \ge \theta \}$ denoted the timing of the first spike in the V1 layer.
Membrane potentials were calculated up to this time point, after which a \ac{WTAI} scheme \cite{maass2000computational} was triggered and all membrane potentials were reset to zero.
In this scheme, the most frequently firing neuron exerted the strongest inhibition on its competitors and thereby stopped them from firing until the end of the iteration.

\subsection{Spike-latency code}

\noindent Following \cite{chauhan2018emergence}, we converted the LGN activity maps to first-spike relative latencies using a simple inverse operation: $y = 1/x$, where $x$ was the LGN input and $y$ was the assigned spike-time latency \cite{masquelier2007unsupervised}. In this way, we ensured that the most active units fired first, while units with lower activity fired later or not at all.

\subsection{Spike-timing dependent plasticity}

\noindent The weights of plastic synapses connecting \ac{LGN} and V1 were updated using \ac{STDP}, which is an unsupervised learning rule that modifies synaptic strength, $w$, as a function of the relative timing of pre- and postsynaptic spikes, $\Delta t$ \cite{gutig_learning_2003}. \Ac{LTP} ($\Delta{t} > 0$) and \ac{LTD} ($\Delta{t} \le 0$) were driven by their respective learning rates $\alpha^{+}$ and $\alpha^{-}$, leading to a weight change ($\Delta w$):
\begin{equation}
\Delta{w}=\begin{cases} -\alpha^{-} \cdot w^{\mu^{-}} \cdot K(\Delta{t},\tau_{-}), \Delta{t} \le 0 \\
                     \alpha^{+} \cdot (1-w)^{\mu^{+}} \cdot K(\Delta{t},\tau_{+}), \Delta{t} > 0,
       \end{cases} \label{eq:1}
\end{equation}
where $\alpha^{+} = 5 \times 10^{-3}$ and $\alpha^{-} = 3.75 \times 10^{-3}$, $K(\Delta{t}, \tau) = e^{-\vert\Delta{t}\vert/\tau}$ was a temporal windowing filter, and $\mu^{+} = 0.65$ and $\mu^{-} = 0.05)$ were constants $\in [0, 1]$ that defined the nonlinearity of the \ac{LTP} and \ac{LTD} process, respectively. 
In this implementation, computation speed greatly increased by making the windowing filter $K$ infinitely wide (equivalent to assuming $\tau_{\pm} \to \infty$, or $K = 1$) \cite{gutig_learning_2003}.
% For more details see \cite{chauhan2018emergence}.

A ratio $\alpha^{+} /\alpha^{-} = 4/3$ was chosen based on previous experiments that demonstrated network stability \cite{masquelier2007unsupervised}.
The threshold of the V1 neurons was fixed through trial and error at $\theta = 20$. This value was unmodified for all experiments.

Initial weight values were sampled from a random uniform distribution between 0 and 1.
After each iteration, the synaptic weights for the first V1 neuron to fire were updated using STDP (Equation~\ref{eq:1}), and the membrane potentials of all the other neurons in the V1 population were reset to zero. 
The \ac{STDP} rule was active only during the training phase.
% and the synaptic weights for the first V1 neuron to fire were updated using the Equation~\ref{eq:1}. 
% When a neuron was repeatedly presented with similar inputs, 
\ac{STDP} has the effect of concentrating high synaptic weights on afferents that systematically fire early, thereby decreasing postsynaptic spike latencies for these connections.
% This leads to an efficient population code, where the probability of any two neurons learning the same feature is highly reduced.

\subsection{Winner-take-all inhibition}

\noindent We used a hard \ac{WTAI} scheme such that, if any V1 neuron fired during a certain iteration, it simultaneously prevented other neurons from firing until the next sample \cite{maass2000computational}. This scheme computes a function \ac{WTAI}$_n$$:\mathbb{R}^{n}\rightarrow \lbrace0, 1\rbrace^n$  whose output $\langle y_1,..., y_n \rangle $ =  \ac{WTAI}$_n$ ($x_1$,..., $x_n$) satisfied:

\begin{equation}
y_i =\begin{cases} 1, \,\,\,\, $if  $x_i$ $>$ $x_j$ for all $j \neq i$ $\\
                     0, \,\,\,\, $if  $x_j$ $>$ $x_i$ for some $j \neq i$ $. 
       \end{cases} \label{eq:3}
\end{equation}

For a given set of $n$ different inputs $x_{1},..., x_{n}$, a hard \ac{WTAI} scheme would thus yield a single output $y_{i}$ with value $1$ (corresponding to the neuron that received the largest input $x_{i}$), whereas all other neurons would be silent. We also implemented various soft WTA-I schemes to investigate how the quality of the represented objects changes. The soft WTA-I schemes consisted of 10, 50, 100 and 150 (i.e., all V1 neurons) neurons firing during a certain iteration, while all other neurons were silenced (see Figure~\ref{fig:graph2}).

\begin{figure}[!t]
     \centering
     \begin{subfigure}[b]{0.07\textwidth}
         \centering
         \footnotesize Original \\
         \includegraphics[width=\textwidth]{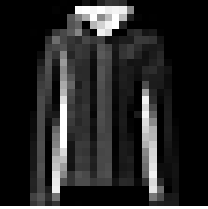}
     \end{subfigure}
     \begin{subfigure}[b]{0.07\textwidth}
         \centering
         \footnotesize LGN \\
         \includegraphics[width=\textwidth]{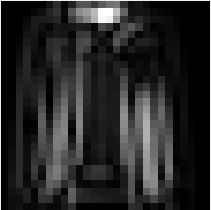}
     \end{subfigure}
     \begin{subfigure}[b]{0.07\textwidth}
         \centering
          \footnotesize OR \\
         \includegraphics[width=\textwidth]{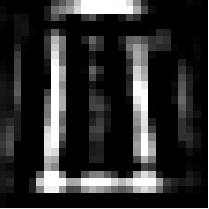}
     \end{subfigure}
     \begin{subfigure}[b]{0.07\textwidth}
         \centering
         \footnotesize Original \\
         \includegraphics[width=\textwidth]{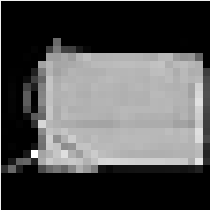}
     \end{subfigure}
     \begin{subfigure}[b]{0.07\textwidth}
         \centering
         \footnotesize LGN \\
         \includegraphics[width=\textwidth]{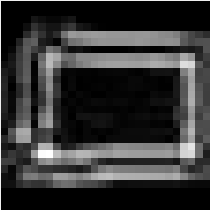}
     \end{subfigure}
     \begin{subfigure}[b]{0.07\textwidth}
         \centering
         \footnotesize OR \\
         \includegraphics[width=\textwidth]{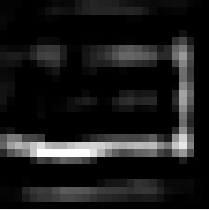}
     \end{subfigure}
     \begin{subfigure}[b]{0.07\textwidth}
         \centering
         \includegraphics[width=\textwidth]{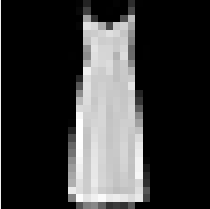}
     \end{subfigure}
     \begin{subfigure}[b]{0.07\textwidth}
         \centering
         \includegraphics[width=\textwidth]{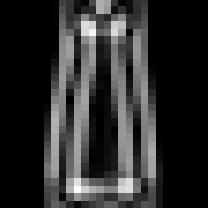}
     \end{subfigure}
     \begin{subfigure}[b]{0.07\textwidth}
         \centering
         \includegraphics[width=\textwidth]{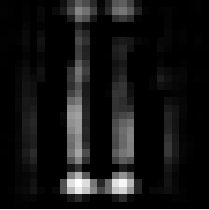}
     \end{subfigure}
          \begin{subfigure}[b]{0.07\textwidth}
         \centering
         \includegraphics[width=\textwidth]{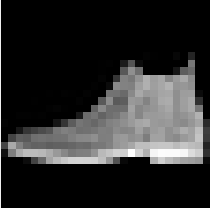}
     \end{subfigure}
     \begin{subfigure}[b]{0.07\textwidth}
         \centering
         \includegraphics[width=\textwidth]{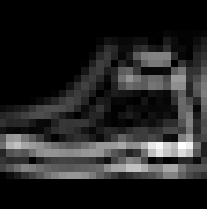}
     \end{subfigure}
     \begin{subfigure}[b]{0.07\textwidth}
         \centering
         \includegraphics[width=\textwidth]{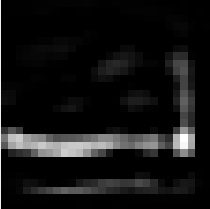}
     \end{subfigure}
          \begin{subfigure}[b]{0.07\textwidth}
         \centering
         \includegraphics[width=\textwidth]{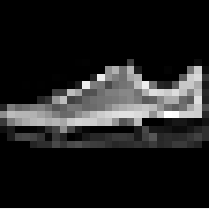}
     \end{subfigure}
     \begin{subfigure}[b]{0.07\textwidth}
         \centering
         \includegraphics[width=\textwidth]{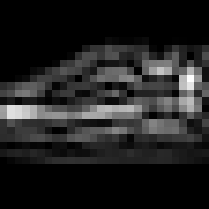}
     \end{subfigure}
     \begin{subfigure}[b]{0.07\textwidth}
         \centering
         \includegraphics[width=\textwidth]{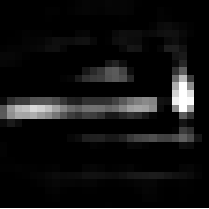}
     \end{subfigure}
          \begin{subfigure}[b]{0.07\textwidth}
         \centering
         \includegraphics[width=\textwidth]{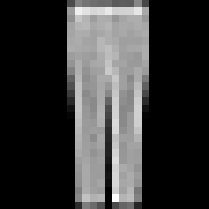}
     \end{subfigure}
     \begin{subfigure}[b]{0.07\textwidth}
         \centering
         \includegraphics[width=\textwidth]{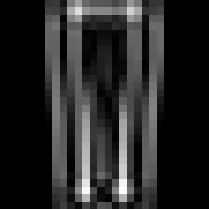}
     \end{subfigure}
     \begin{subfigure}[b]{0.07\textwidth}
         \centering
         \includegraphics[width=\textwidth]{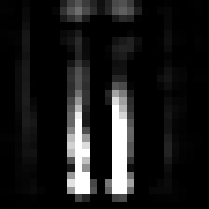}
     \end{subfigure}
        \caption{Examples of object representation. First and fourth columns: Fashion MNIST samples. Second and fifth columns: LGN activity maps (after preprocessing). Third and sixth columns: Object representation (OR) using a hard \ac{WTAI} scheme with 150 V1 neurons, which aimed to reconstruct the LGN activity maps.}
        \label{fig:reconstructedImages}
\end{figure}

\subsection{Dataset}

\noindent We assessed the ability of our SNN network to represent visual stimuli using the Fashion MNIST database \cite{xiao2017fashion}. The Fashion-MNIST dataset comprises $28 \times 28$ grayscale images of 70,000 fashion products from 10 categories, with 7,000 images per category. To train the network, we randomly selected 1,000 training images and 200 test images.

\subsection{Stimulus reconstruction}

\noindent The post-convergence receptive field $\xi_{j}$ of the $i$-th V1 neuron was estimated as follows:
\begin{equation}
\xi_{j} \approx  \sum\limits_{j\in{LGN}}w_{ij}\psi_{j},
\label{eq:2}
\end{equation}
where $\psi_{j}$ was the receptive field of the $j$-th LGN afferent, and $w_{ij}$ was the weight of the synapse connecting the $j$-th afferent to the $i$-th V1 neuron. 

Stimuli $k$ were then linearly reconstructed from the V1 population activity:
% For object representation \textit{(OR)} after the training phase, we used a linear approximation using the activation and receptive fields of each V1 neuron, as follow:
\begin{equation}
OR_{k}=  \sum\limits_{j\in{V1}}r_{kj}\xi_{j},
\label{eq:5}
\end{equation}
where $r_{kj}$ was the response of the $j$-th V1 neuron to the $k$-th image and $\xi_{j}$ was its receptive field.

\section{Results}

\begin{figure}[t]
  \centering
   \includegraphics[width=0.8\linewidth]{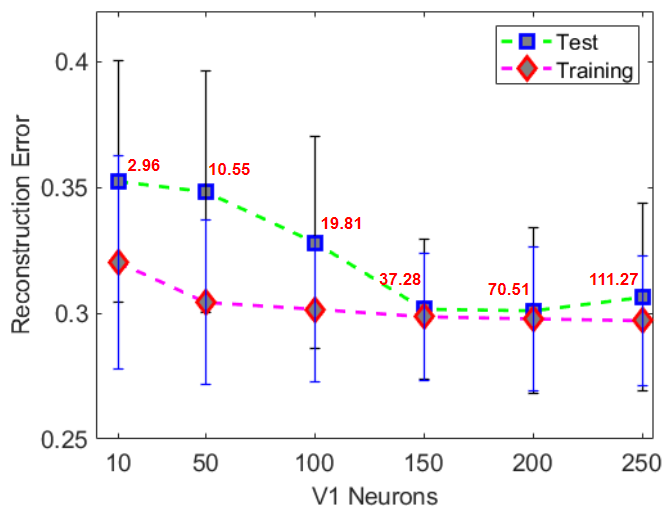}
   \caption{Reconstruction error for the training and test sets using different V1 neurons for a hard \ac{WTAI} scheme (\ac{WTAI} = 1). Number of spikes needed to optimally represent an object during the test phase is given in red.}
   \label{fig:graph1}
\end{figure}

\begin{figure}[t]
     \centering
     \begin{subfigure}[b]{0.06\textwidth}
         \centering
         \scriptsize LGN \\
         \includegraphics[width=\textwidth]{figures/imgRec/lgn1.png}
         \vspace{1.5em}
     \end{subfigure}
     \begin{subfigure}[b]{0.06\textwidth}
         \centering
         \scriptsize V1 = 10 \\
         \includegraphics[width=\textwidth]{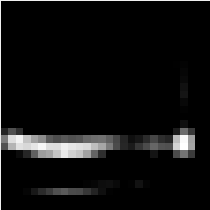}
         \footnotesize 0.3308 \hspace{2em}
     \end{subfigure}
     \begin{subfigure}[b]{0.06\textwidth}
         \centering
         \scriptsize V1 = 50 \\
         \includegraphics[width=\textwidth]{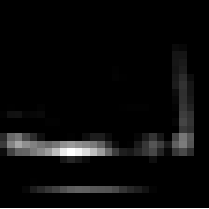}
         \footnotesize 0.3229 \hspace{2em}
     \end{subfigure}
     \begin{subfigure}[b]{0.06\textwidth}
         \centering
         \scriptsize V1 = 100 \\
         \includegraphics[width=\textwidth]{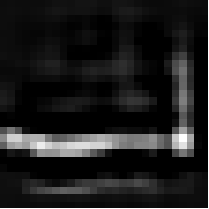}
         \footnotesize 0.3076 \hspace{2em}
     \end{subfigure}
     \begin{subfigure}[b]{0.06\textwidth}
         \centering
         {\color{red}\scriptsize V1 = 150 }\\
         \includegraphics[width=\textwidth]{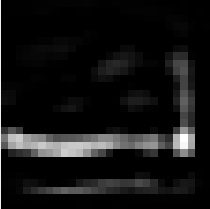}
        \footnotesize 0.2964 \hspace{2em}
     \end{subfigure}
     \begin{subfigure}[b]{0.06\textwidth}
         \centering
         \scriptsize V1 = 200 \\
         \includegraphics[width=\textwidth]{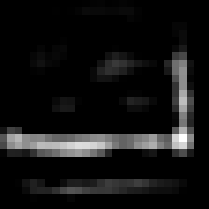}
        \footnotesize 0.2951 \hspace{2em}
     \end{subfigure}
     \begin{subfigure}[b]{0.06\textwidth}
         \centering
         \scriptsize V1 = 250 \\
         \includegraphics[width=\textwidth]{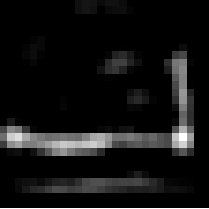}
        \footnotesize 0.3007 \hspace{2em}
     \end{subfigure}
        \caption{Representation of a shoe using a population of V1 neurons (ranging from 10 to 250 neurons). The number below each image indicates the reconstruction error for that particular image.}
        \label{fig:objectrepresentationV1}
\end{figure}

% ----------------------------------------

\subsection{Object representation using hard WTA-I}

\noindent Example object reconstructions obtained after training with a hard \ac{WTAI} scheme (i.e., where only one neuron was active for each training image) are shown in Figure~\ref{fig:reconstructedImages}.
Here, every spiking neuron became selective for a particular object feature (example receptive fields learned by \ac{STDP} are shown in Figure~\ref{fig:approach}), so that after training a whole object could be represented by a linear combination of V1 neurons (Equation~\ref{eq:5}). 

Figure~\ref{fig:graph1} shows the reconstruction error after training for both the training and the test sets using different V1 neurons for a hard \ac{WTAI} scheme.
Reconstruction error for an image $k$ was calculated as the mean square error between the LGN activity map ($\mathrm{LGN}_k$) and $\mathrm{OR}_k$.
Figure~\ref{fig:graph1} reports the mean and standard deviation of all reconstruction errors across the train and test sets, respectively.
We found that the reconstruction error for the training set decreased with an increasing number of V1 neurons.
On the other hand, the reconstruction error of the test set went through a minimum (at roughly 150 V1 neurons), which is consistent with the bias-variance dilemma \cite{beyeler2019neural}.
In addition, the number of spikes needed to represent an object increased with the number of V1 neurons, nearly doubling from 37.28 spikes at 150 neurons to 70.51 spikes at 200 neurons.
Increasing the V1 population beyond 150 neurons did therefore not lead to any visible benefits in reconstruction error (Figure~\ref{fig:objectrepresentationV1}), but required many more spikes to represent an object.

\subsection{Object representation using soft WTA-I schemes}

\begin{figure}[t]
  \centering
   \includegraphics[width=0.7\linewidth]{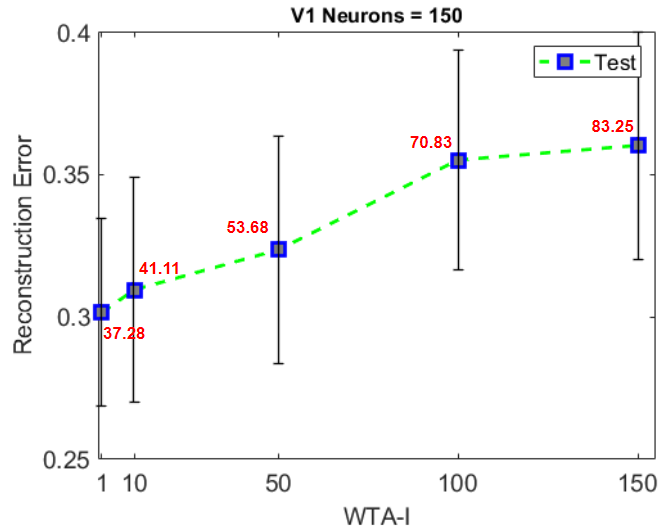}
   \caption{Reconstruction error in the test phase as a function of the number of spikes included in the STDP algorithm (WTA-I) for 150 V1 neurons. Number of spikes needed to optimally represent an object during the test phase is given in red.}
   \label{fig:graph2}
\end{figure}

\begin{figure}[t]
     \centering
      \begin{subfigure}[b]{0.069\textwidth}
         \centering
         \scriptsize LGN \\
         \includegraphics[width=\textwidth]{figures/imgRec/lgn1.png}
         \vspace{0.2em}
     \end{subfigure}
     \begin{subfigure}[b]{0.07\textwidth}
         \centering
         {\color{red}\scriptsize \ac{WTAI}-1} \\
         \includegraphics[width=\textwidth]{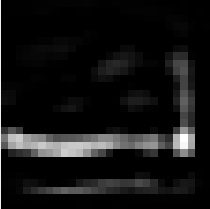}
         \footnotesize 0.2964 
     \end{subfigure}
     \begin{subfigure}[b]{0.07\textwidth}
         \centering
         \scriptsize \ac{WTAI}-10 \\
         \includegraphics[width=\textwidth]{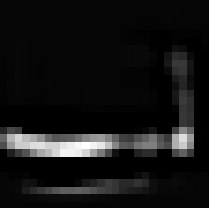}
         \footnotesize	0.3007 
     \end{subfigure}
     \begin{subfigure}[b]{0.07\textwidth}
         \centering
         \scriptsize\ac{WTAI}-50 \\
         \includegraphics[width=\textwidth]{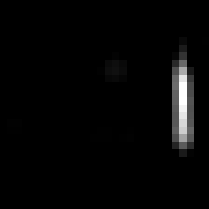}
         \footnotesize	0.3164 
     \end{subfigure}
     \begin{subfigure}[b]{0.07\textwidth}
         \centering
         \scriptsize \ac{WTAI}-100 \\
         \includegraphics[width=\textwidth]{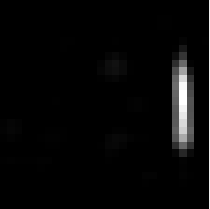}
        \footnotesize 0.3415 
     \end{subfigure}
     \begin{subfigure}[b]{0.07\textwidth}
         \centering
        \scriptsize \ac{WTAI}-150 \\
         \includegraphics[width=\textwidth]{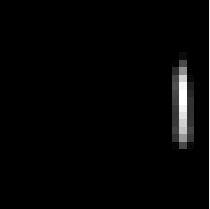}
        \footnotesize 0.3508 
     \end{subfigure}
        \caption{Representation of a shoe using different \ac{WTAI} schemes, where between 1 (WTA-I 1) and 150 (WTA-I 150) neurons were active for each training sample. The number below each image indicates the reconstruction error for that particular image.}
        \label{fig:WTAI}
\end{figure}

\noindent We also tested object representation using various soft \ac{WTAI} schemes, where we varied the number of V1 neurons allowed to be active for each training image.
Figure~\ref{fig:graph2} shows the reconstruction error on the test set across the range of possible \ac{WTAI} schemes, ranging from hard (where for every image only one neuron was active) to soft (where all neurons (150) were active).
We found that the softer the \ac{WTAI} scheme, the higher the reconstruction error and the number of spikes needed to represent an object.
The reason for this became evident when we visualized the resulting object representations (Figure~\ref{fig:WTAI}).
\Ac{WTAI} schemes where at most 10 neurons were allowed to be active where instrumental in maintaining competition among neurons. In the absence of a strong \ac{WTAI} scheme, multiple neurons ended up learning similar visual features, which resulted in poor object reconstructions (right half of Figure~\ref{fig:WTAI}).

\section{Conclusion}
\noindent We have shown that a network of spiking neurons relying on biologically plausible learning rules and coding schemes can efficiently represent objects from the Fashion MNIST dataset with as little as 40 spikes.
\Ac{WTAI} schemes were essential for enforcing competition among neurons, which led to sparser object representations and lower reconstruction errors.
A future extension of the model might focus on deeper architectures and more challenging visual stimuli.

{\small
\bibliographystyle{ieee_fullname}
\bibliography{egbib}
}

\end{document}